\documentclass{article}

\usepackage[preprint]{neurips_2023}

\usepackage[utf8]{inputenc} 
\usepackage[T1]{fontenc}    
\usepackage{url}            
\usepackage{booktabs}       
\usepackage{amsfonts}       
\usepackage{nicefrac}       
\usepackage{microtype}      
\usepackage{xcolor}         
\usepackage{graphicx}
\usepackage{booktabs}       
\usepackage{multirow}
\usepackage{makecell}
\usepackage{amsmath}
\usepackage{colortbl}
\usepackage{color}

\usepackage{natbib}
\setcitestyle{numbers,square}

\usepackage[colorlinks,
            linkcolor=red,
            anchorcolor=blue,
            citecolor=green
            ]{hyperref}

\definecolor{darkgreen}{RGB}{30,130,30} 
\title{DiT: Efficient Vision Transformers with Dynamic Token Routing}

\author{%
  {Yuchen Ma}, {Zhengcong Fei}, {Junshi Huang} \\
  Meituan \\
  \small{\{mayuchen, feizhengcong, huangjunshi\}@meituan.com}
}

\begin{document}

\maketitle

\begin{abstract}
Recently, the tokens of images share the same static data flow in many dense networks. However, challenges arise from the variance among the objects in images, such as large variations in the spatial scale and difficulties of recognition for visual entities. In this paper, we propose a data-dependent token routing strategy to elaborate the routing paths of image tokens for Dynamic Vision Transformer, dubbed DiT. The proposed framework generates a data-dependent path per token, adapting to the object scales and visual discrimination of tokens. In feed-forward, the differentiable routing gates are designed to select the scaling paths and feature transformation paths for image tokens, leading to multi-path feature propagation. In this way, the impact of object scales and visual discrimination of image representation can be carefully tuned. Moreover, the computational cost can be further reduced by giving budget constraints to the routing gate and early-stopping of feature extraction.
In experiments, our DiT achieves superior performance and favorable complexity/accuracy trade-offs than many SoTA methods on ImageNet classification, object detection, instance segmentation, and semantic segmentation. Particularly, the DiT-B5 obtains 84.8\% top-1 Acc on ImageNet with 10.3 GFLOPs, which is 1.0\% higher than that of the SoTA method with similar computational complexity. These extensive results demonstrate that DiT can serve as versatile backbones for various vision tasks. The code is available at \href{https://github.com/Maycbj/DiT}{https://github.com/Maycbj/DiT}.

\end{abstract}

\section{Introduction}
\begin{figure}[tbp]
\centering
\includegraphics[clip=true, width=1.0\linewidth]{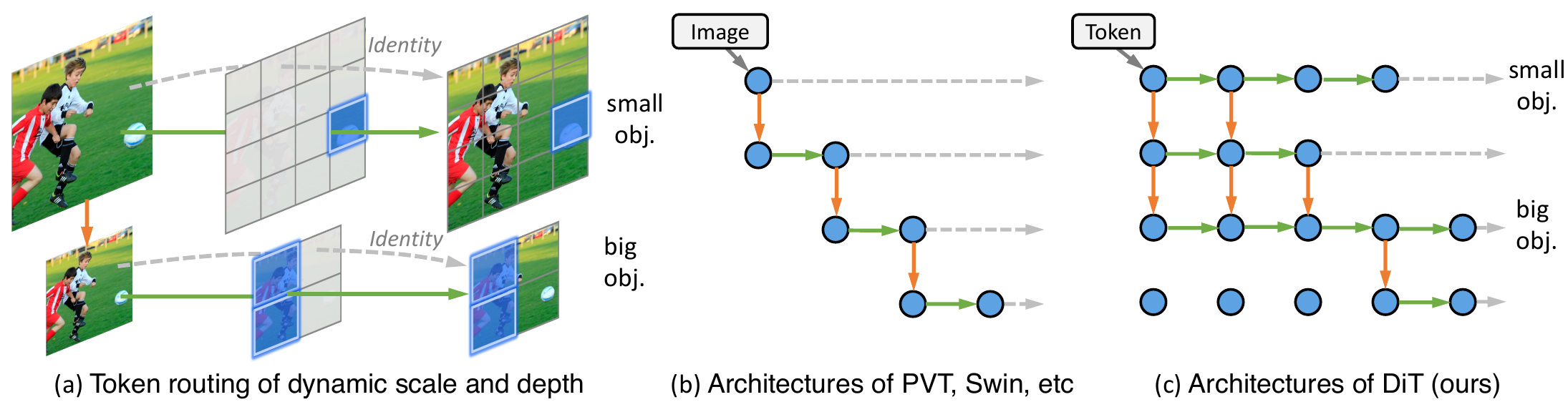}
\caption{In all figures, dotted grey lines represent the identical transformation of token features, solid green lines represent the feature extraction by transform block, and solid orange lines represent the down-scaling of tokens.
\textbf{(a) Illustration of our main idea.} Given input image with different spatial scales, the proposed \textit{dynamic token routing} adaptively selects the \textit{\textbf{token-specific}} forward paths, including dynamic-scaling and dynamic-depth paths. 
In this way, scale-variant objects (\emph{e.g.} football and children) may be activated on feature maps with appropriate resolutions. 
Meanwhile, the fine-grained tokens (\emph{e.g.} children) require more elaborated procedures of feature extraction.
\textbf{(b)(c)} present the difference between existing methods and our proposed DiT, where the node represents the image/token feature.
Please note that the dynamic path in DiT is selected in a token-specific manner, comparing with image-specific configuration in PVT, Swin, and stochastic depth networks.
} 
\label{fig:preshow}
\end{figure}

\vspace{-4pt}
Over the past few years, transformer-based networks have achieved promising results in various computer vision tasks, including image classification~\cite{vit,swin}, object detection~\cite{detr,transformer_det}, and semantic segmentation~\cite{segformer}. 
However, some problems in those networks are still unresolved, one of which concentrates on the variance of object representations mainly stemming from two aspects: 
(1) The variance of object scales in the image. An example is presented in Fig.~\ref{fig:preshow}(a), where objects of different scales indicate the diverse number of tokens with different appearances of details.
(2) The variance of feature discrimination for coarse-grained and fine-grained object recognition. For example, the recognition of ``human'' object may require higher-level semantic features than ``football'' object and background.

The large variance in token appearance and feature discrimination brings difficulties to feature representation learning, traditional methods try to solve this problem by well-designed network architectures. 
Conventional CNNs~\cite{Inception,resnet}, as well as recently developed transformer-based networks~\cite{pvt,swin,crossvit} usually follow a sequential methodology in their architecture design, which involves a gradual reduction in the spatial size of feature maps. 
This paradigm leads to a progressive shrinking pyramid to highlight the representation of large and discriminative objects while neglecting the unimportant regions.
For dense prediction tasks like object detection and instance segmentation, multiple-scale features are indispensable for handling objects of varying scales.
The canonical approach for multi-scale features involves FPN from RetinaNet~\cite{retinanet} and Mask R-CNN~\cite{maskrcnn}, where the assignment of ground-truth labels in feature pyramid greatly depends on the scales of objects (\emph{e.g.} the prediction of small objects occurs at the high resolution of feature map). However, those handcrafted architectures fully utilize the multi-scale features in a static way, we propose to select some appropriate scaling paths for image tokens as dynamic-scaling networks.

For the comprehensive learning of image representation, many recent works~\cite{resnet,deeper} directly use some deeper and more complicated backbones~\cite{coca,crossvit,cfnet}.
Particularly, some works~\cite{rao2021dynamicvit,ats} propose dynamic token adaptation strategies to discard or merge redundant tokens in classification tasks, leading to promising accuracy and Flops trade-off. 
However, these pruning-like strategies are hardly applied to the dense prediction tasks, due to the scarcity of pruned image tokens and unstructured pruned output.
To eliminate this problem, we propose to identically transform image tokens in some blocks via skip-path, rather than token pruning.
Consequently, the discrimination of token features from variant objects or background can be comprehensively learned, where the network tends to skip the feature extraction of simple objects or background, and requires more procedures of feature extraction for complex objects.
With the learnable routing strategy, \textbf{the tokens in single image} conduct different feature transformation paths, describing the main difference between our dynamic-depth network to stochastic depth networks~\cite{stochastic}.

In this paper, we integrate a dynamic-scaling network and dynamic-depth network into vision transformer, composing the main structure of Dynamic-Vision-Transformer, dubbed DiT. 
As presented in Fig.~\ref{fig:preshow}(c), we construct the grid-like network structure, and each node represents the \emph{token routing module}.
The token routing module determines the calculated path of each token and predicts the probability distribution of whether to be calculated by the transformer module. We hope to explore the unstructured and data-dependent token routing strategy for vision transformers.
In the routing module, two binary gates are learned for the path selection of each token, where the vertical and horizontal paths correspond to the \textbf{dynamic-scaling} and \textbf{dynamic-depth} path, respectively.
In this manner, response maps from proper sizes and depths of layers are dynamically activated for variant scales and the complexity of token paths. 
To balance the trade-off between performance and Flops, the constraints of computational budgets (\emph{e.g.} FLOPS) are considered in our learning task.
As our designed differentiable routing gates and computational constraints, the DiT could be learned in end-to-end manner with dynamic path selection per token on the fly. 
The overall approach can be easily integrated into most transformer-based vision backbones. 

To elaborate on its superiority in both performance and efficiency, we follow the standard training recipe as in DeiT~\cite{deit}, and consistently achieve superior performance compared to recent SOTA Vision Transformers~\cite{vit,swin,pvt,mpvit}. 
Furthermore, We integrate DiT as the backbone into dense prediction tasks, \emph{e.g.} object detection and instance segmentation task, on the COCO dataset and semantic segmentation on the ADE20K dataset, achieving state-of-the-art performance on all benchmarks.

\section{Related Works}
\vspace{-4pt}
\paragraph{Transformer backbones.} 
ViT~\cite{vit} is the pioneer work to demonstrate that a pure Transformer can achieve state-of-the-art performance in image classification. ViT employs a pure Transformer model for image classification by treating an image as a sequence of patches. DeiT~\cite{deit} further explores a data-efficient training strategy and a distillation method for ViT. After that, more recent methods such as CPVT~\cite{cpvt}, TNT~\cite{tnt}, and CrossViT~\cite{crossvit} further improve the performance of ViT by modifying the network structure.

Beyond classification, PVT ~\cite{pvt} is an important work to introduce a pyramid structure in Transformer, which proves the potential of a pure Transformer backbone compared to CNN counterparts in dense prediction tasks. Subsequently, methods such as Swin~\cite{swin}, CvT~\cite{cvt}, CoaT~\cite{CoaT}, and LeViT~\cite{levit} enhance the local continuity of features and show the flexibility to model at various scales with linear computational complexity with respect to image size.

\vspace{-2pt}
\paragraph{Multi-scale feature for Dense prediction.} 
For dense prediction tasks, One of the problems in dense prediction comes from the huge scale variance among inputs, e.g., the tiny object instances and the picture-filled background stuff. 
Therefore, most existing works combine an image classification network and a feature fusion module to extract and fuse multi-scale features. 
FPN~\cite{fpn} uses a top-down pathway to sequentially combine multi-level features. 
Subsequently, the label assignment of the multi-scale feature pyramid in RetinaNet~\cite{retinanet} and Mask R-CNN~\cite{maskrcnn} depends on the scale of the objects. PANet~\cite{panet} further introduces a bottom-up pathway to shorten the pathway between features. DeeplabV3+~\cite{deeplabv3} fuses multi-scale features obtained by atrous spatial pyramid pooling. 
MPViT~\cite{mpvit} exploits tokens of different scales independently fed into the Transformer encoders via multiple paths and the resulting features are aggregated. Overall, those handcrafted architectures aim at utilizing multi-scale features in a static network, rather than adapting to input dynamically.

\vspace{-2pt}
\paragraph{Dynamic Networks.} 
Dynamic networks, adaptively adjusting the network architecture to the corresponding input, have been recently studied in the current network design. For the traditional CNN backbones, previous methods mainly focus on image classification by dropping blocks ~\cite{wu2018blockdrop,huang2017multi} or pruning channels ~\cite{Lin2017RuntimeNP,you2019gate} for efficient inference. ~\cite{li2020learning} propose a framework generating data-dependent routes, adapting to the scale distribution of each image on the CNN architecture. Recently, many works explore the effects of dynamic networks in transformer-based vision backbones. 
DynamicVit~\cite{rao2021dynamicvit}, a dynamic token sparsification framework to prune redundant tokens progressively and dynamically based on the input, which gradually drops the uninformative tokens as the computation proceeds on the image classification task. ~\cite{ats} introduce a differentiable parameter-free Adaptive Token Sampling module, which can be plugged into any existing vision transformer architecture. However, all these works focus on transformer-based token pruning, and few works explore the possibility of making use of the characteristic of token routing to accelerate the dense prediction tasks. To utilize the dynamic property, an end-to-end dynamic routing framework is proposed in this paper to learn multi-scale token routing and improve the performance of the downstream tasks.


\section{Dynamic Token Routing}
Compared with previous static network architectures, \textit{dynamic token routing} has the superiority in better performance with the budgeted resource consumption. In this section, we first introduce the architecture of the DiT. Secondly, the details of the routing gate module will be clarified. We illustrate the constrained budgets and training details at the end of this section. 

\subsection{Overview Architecture}
\begin{figure}[tbp]
\vspace{-6pt}
\centering
\includegraphics[clip=true, width=1.0\linewidth]{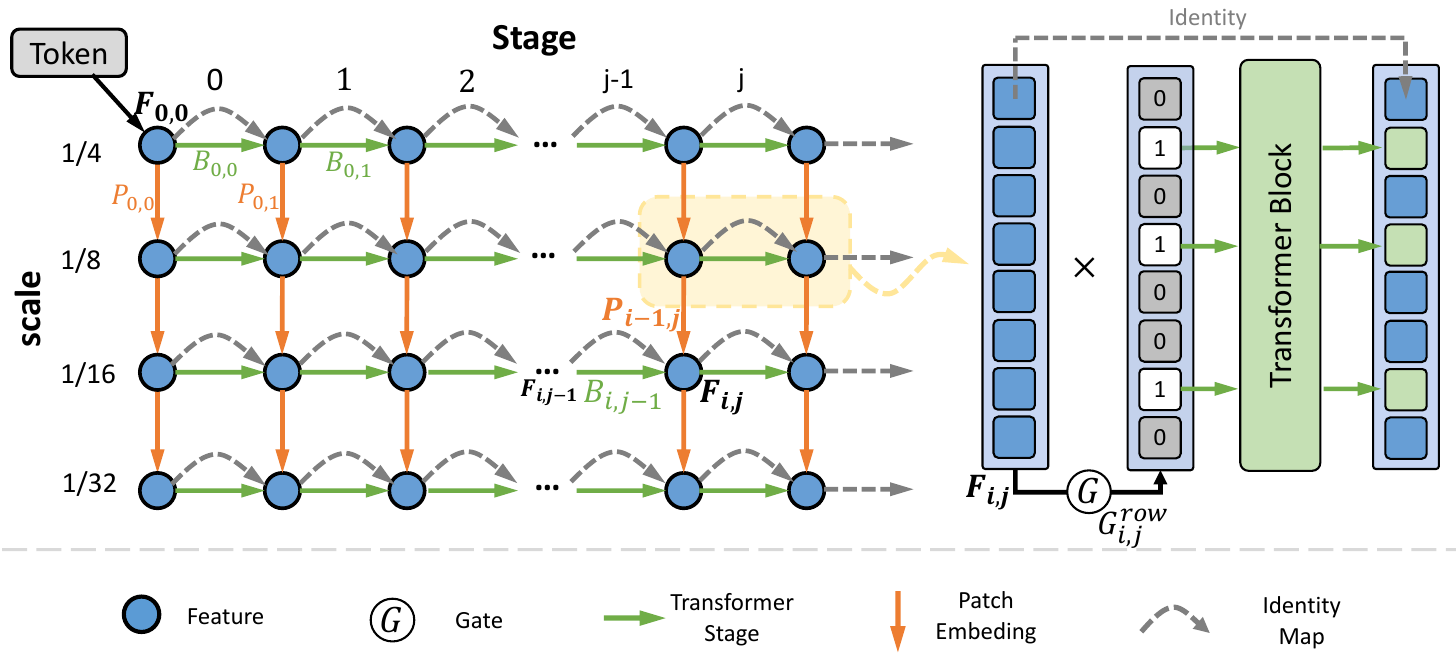}
\caption{The proposed \textit{dynamic token routing} framework of DiT. Left: The routing space with layer \textit{L} and max downsampling rate 1/32. Lines in green, orange, and gray are alternative paths for dynamic routing, which represent Transformer Block, Patch Embedding Block, and identity mapping layer, respectively. Given the feature map from the former layer, we first generate a mask using the \textit{routing gate}. The discrete value of the mask indicates whether the token is involved in the calculation of this layer. More details about the network are elaborated in Sec.~\ref{routing_prediction_modules} Best viewed in color. }
\label{fig:method}
\end{figure}

\vspace{-3pt}
Our main idea is to integrate the dynamic token routing strategy into the transformer-based network to dynamically select the token-specific forward path through multi-scale feature maps and various network depths. Fig.~\ref{fig:method} presents the overview of DiT, where we follow the similar principle of network design as PVT~\cite{vit}. 
Specifically, we construct our network as a grid-like structure, and each node in the grid represents the response map used as the input of the following layers. 
For example, the response map at $(i,j)$ stage of the grid is denoted as $F_{i,j}$, which is then used as input of patch embedding layer $\mathcal{P}_{i,j}$, transformer blocks $\mathcal{B}_{i,j}$, and identity mapping layer.
Note that the token routing strategy is also dependent on this feature map, more details will be introduced in Sec.~\ref{routing_prediction_modules}.

Basically, the token features of response map $F_{i,j}$ are calculated by element-wise sum with routing strategy on 3 types of inputs at most, \emph{i.e.}, the output of transformer blocks $\mathcal{B}_{i, j-1}$, the output of patch embedding layer $\mathcal{P}_{i-1, j}$, and the identity mapping of $F_{i, j-1}$.
Note that the calculations of some token features in the transformer blocks and patch embedding layer may be skipped due to routing strategy, we simply mask the corresponding features of response maps in such cases.

In our network, the input image with dimensions $H \times W \times 3$ is divided into $HW/4^{2}$ patches in the first stage, which reduces the resolution to 1/4 of the raw size. 
After that, the space composed of patch embedding layers, transformer stages, and identity mapping layers are designed for dynamic routing, called \textit{token routing space}. 
Generally, the entire space has $N$ levels of patch embedding layers, each of which downsamples the scale of input maps into 1/2.
Usually, we set $N$ as 4 in our experiments, leading to 1/32 downsampling of the original image as presented in Fig.~\ref{fig:method}.
In the same level of patch embedding, the number of transformer stages is set as $M$.
For a fair comparison, the stage number $M$ and the number of transformer blocks in each stage are similar to that of PVT~\cite{pvtv2}.




\vspace{-3pt}
\subsection{Dynamic Token Routing}
\label{routing_prediction_modules}
An important characteristic of our DiT is that the token routing is performed in a self-dependent manner, \emph{i.e.}, the token features in response map $F_{i,j}$ are adopted to control the routing paths of themselves. Generally, both the transformer stage and patch embedding module have a binary gate to select the token routing paths, respectively.

For the routing gate in each transformer stage, a binary decision mask $G \in \{0, 1\}^{l}$ is used to decide whether to calculate the token features via transformer blocks, where $l = h \times w $ is the number of tokens at the corresponding stage.
Initially, we set all elements of the decision mask as 1. 
To learn the data-dependent mask at stage $(i,j)$ of grid-like network , we use a linear projection $w_{i,j}^{row} \in \mathbb{R}^{d \times 2}$ to predict the probability map $P_{i,j}^{row}$ according to the feature map $F_{i,j} \in \mathbb{R}^{l \times d}$, where $d$ is the feature dimension: 
\begin{equation}\label{eq:probabilities}
P_{i,j}^{row} = Softmax(F_{i,j} * w_{i,j}^{row}) \in \mathbb{R}^{l \times 2}.
\end{equation}
\vspace{-6pt}

As our target is to perform token-specific routing, we need to discrete the probability map of tokens.
Therefore, the Gumbel-Softmax~\cite{gumbel} is used to sample the binary decision mask from the probability map $P_{i,j}^{row}$ as follows: 
\vspace{-6pt}
\begin{equation}\label{eq:gumbel}
G_{i,j}^{row} = Gumbel\text{-}Softmax(P_{i,j}^{row}) \in \{0,1\}^{l}
\end{equation}
\vspace{-12pt}

In order to obtain a more stable gating module in inference, we maintain a moving averaged version of linear projection weight $\hat{w}_{i,j}^{row}$ by momentum update: 
\begin{equation}\label{eq:momentum}
\hat{w}_{i,j}^{row} \leftarrow m * \hat{w}_{i,j}^{row} + (1 - m) * w_{i,j}^{row},
\end{equation}
where $m \in [0, 1)$ is a momentum coefficient, and $w_{i,j}^{row}$ are updated by back-propagation. 

The routing gate in patch embedding layer decides whether to downsample the response map or not. 
Similar to the routing gate in the transformer stage, we generate a binary scalar $G_{i,j}^{col}$ for the routing decision of the entire feature map: 
\begin{equation*}
\label{eq:probabilities_col}
\begin{aligned}
P_{i,j}^{col} &= Softmax(\bar F_{i,j} * w_{i,j}^{col}) \in \mathbb{R}^{1 \times 2} \\
G_{i,j}^{col} &= Gumbel\text{-}Softmax(P_{i,j}^{col}) \in \{0,1\}
\end{aligned}
\end{equation*}
where $\bar F_{i,j}\in \mathbb{R}^{d}$ is obtained by average pooling on the response map $F_{i,j}$. 
Similarly, the moving averaged weight $\hat w_{i,j}^{col}$ is maintained in the training stage for stable inference.

With the proposed gating module as aforementioned, the feature map $F_{i,j}$ can be formulated as 
\begin{equation}
\label{eq:feature}
\begin{split}
F_{i,j} = \underbrace{\mathcal{B}_{i,j-1}(G_{i,j-1}^{row} \odot F_{i,j-1}) + (1-G_{i,j}^{row}) \odot F_{i,j-1}}_{row} + 
\underbrace{G_{i-1,j}^{col} \odot \mathcal{P}_{i-1,j}(F_{i-1,j})}_{col}
\in \mathbb{R}^{l \times d},
\end{split}
\end{equation}
where $\mathcal{B}_{i,j-1}$ and $\mathcal{P}_{i-1,j}$ are the transformer blocks and patch embedding layer, respectively.
It should be noted that we skip the calculation of token features in transformer blocks $\mathcal{B}_{i,j-1}$ if the corresponding elements of mask map $G_{i,j-1}^{row}$ are 0 in our implementation.

\vspace{-3pt}
\subsection{Complexity Constraint}
To balance the trade-off between efficiency and effectiveness, we add the complexity constraint for dynamic token routing. 
The gate modules are inclined to choose a more efficient routing path instead of calculating some uninformative tokens. We denote constant value $C$ as the computational costs associated with the predefined network complexity, \emph{e.g.}, FLOPs. We formulate the cost of each stage as follows. 
\begin{equation}\begin{split}\label{eq:cost_node}
C_{i,j} = \bar G_{i,j}^{row} \cdot {C}_{i,j}^{row} + G_{i,j}^{col} \cdot {C}_{i,j}^{col} + {C}_{i,j}^{gates}
\end{split}\end{equation}
where ${C}_{i,j}^{row}$, ${C}_{i,j}^{col}$, and $ {C}_{i,j}^{gates}$ are the FLOPs of modules $\mathcal{B}_{i,j}$, $\mathcal{P}_{i,j}$, and routing gates, respectively. 
${\bar G}_{i,j}^{row}$ is the average of binary decision map.
Generally, the complexity constraint of each stage contains the cost of transformer blocks, patch embeddings, and routing gates. $\mathcal{C}^{space}$ and $\mathcal{C}^{base}$ denote the cost of our routing space and the cost of the baseline(PVTv2).
Finally, the complexity constraint of the overall network is designed as:
\vspace{-3pt}
\begin{equation}\begin{split}\label{eq:cost_al}
\mathcal{L}_{C} &= ({C}^{space} / {C}^{base} - \mu )^2 \\
&= (\sum_{i,j\in \mathcal{S}} {C}_{i,j}) / {C}^{base} - \mu)^2
\end{split}\end{equation}
where $\mu$ is the ratio of limiting the computational complexity for the network. 
With different $\mu$, the selected routes in each propagation would be adaptively restricted to corresponding calculated budgets. 
Overall,  the network weights, as well as the routing gates, can be optimized with a joint loss function $\mathcal{L}$ in a unified framework. $\mathcal{L}_{N}$ and $\mathcal{L}_{C}$ denote the loss function of the whole network and complexity constraint:
\vspace{-3pt}
\begin{equation}\begin{split}\label{eq:loss_all}
\mathcal{L} = \mathcal{L}_{N} + \lambda_{\mathcal{C}} \mathcal{L}_{C}
\end{split}\end{equation}
\vspace{-3pt}
where $\lambda_{\mathcal{C}}$ is utilized to balance the optimization of network prediction and complexity cost.

\section{Experiment}
In this section, we illustrate the effectiveness and the efficiency of DiT as a vision backbone on image classification (ImageNet-1K), dense predictions such as object detection and instance segmentation (COCO), and semantic segmentation (ADE20K).

\vspace{-3pt}
\subsection{Image classification on ImageNet-1K}
The experimental settings involve performing image classification experiments on the ImageNet 2012 dataset ~\cite{imagenet}, which contains 1.28 million training images and 50K validation images distributed across 1,000 categories. 
To make a fair comparison, all models are trained on the training set and evaluated with the top-1 error on the validation set.
We strictly follow the setting of DeiT ~\cite{deit}, which employs random cropping, random horizontal flipping, label-smoothing regularization~\cite{reth-inception}, mixup~\cite{mixup}, CutMix~\cite{cutmix} as data augmentations during training. Following the network structure and hyperparameters of different sizes of the PVTv2 range from B1 to B5, the size of the DiT is denoted as DiT-B1 to DiT-B5.

We utilize AdamW~\cite{adamw} with a momentum of 0.9, a mini-batch size of 128, and a weight decay of $5 \times 10^{-2}$ to optimize models. 
The initial learning rate is set to $1 \times 10^{-3}$ and decreased following the cosine schedule~\cite{cosine}. 
All models are trained from scratch for 300 epochs on 8 V100 GPUs. 
For benchmark testing, we use a center crop of the validation set, where a $ 224 \times 224 $ patch is cropped to evaluate the classification accuracy.

In Tab.~\ref{tab:imagenet}, we summarize the results for baseline models and the current SoTA models on the image classification task. We can find our DiT consistently outperforms other methods with similar computational complexity (GFLOPs). For the GFLOPs of DiT scaling from 2.0 to 10.3, our model outperforms the state-of-the-art models consistently with a gap of 1.0 points. It should be noted that the FLOPS of DiT is the average value of multiple experiments. With the model size increases, our Focal-Base model achieves 84.8\%, surpassing all other models using comparable FLOPs.

\begin{table}[t]
\setlength{\tabcolsep}{0.3mm}
\begin{minipage}{0.42\linewidth}
\centering
\caption{Image classification performance on the ImageNet validation set. ``\#Param'' denotes the quantity of valid parameters utilized. }
\label{tab:imagenet}
\scalebox{0.84}{
\begin{tabular}{l|cc|c}
Model & \#Params. & GFLOPs & Top-1(\%) \\
\Xhline{0.5pt}
ResNet18~\cite{resnet} & 11.7 & 1.8 & 69.8\\
DeiT-Tiny/16~\cite{deit} & 5.7 & 1.3 & 72.2\\
PVTv1-Tiny~\cite{pvt} & 13.2 & 1.9 & 75.1 \\
PVTv2-B1~\cite{pvtv2} & 13.1 & 2.1 & 78.7 \\
\rowcolor{gray!20}DiT-B1(ours) & 30.3 & 2.0 & \textbf{79.9} \\
\Xhline{0.5pt}
DeiT-Small~\cite{deit} & 22.1 & 4.6 & 79.9 \\
TNT-S~\cite{tnt} & 23.8 & 5.2 & 81.3 \\
Swin-T~\cite{swin} & 29.0 & 4.5 & 81.3 \\
CvT-13~\cite{cvt} & 20.0 & 4.5 & 81.6 \\
PVTv2-B2~\cite{pvtv2} & 25.4 & 4.0 & 82.0 \\
\rowcolor{gray!20}DiT-B2(ours) & 55.9& 3.8 & \textbf{83.1}  \\
\Xhline{0.5pt}
ResNet101~\cite{resnet} & 44.7 & 7.9 & 77.4 \\
CvT-21~\cite{cvt} & 32.0 & 7.1 & 82.5 \\
PVTv2-B3~\cite{pvtv2} & 45.2 & 6.9 & 83.2 \\
\rowcolor{gray!20}DiT-B3(ours) & 86.3 & 6.8 & \textbf{84.2} \\
\Xhline{0.5pt}
PVTv1-Large~\cite{pvt} & 61.4 & 9.8 & 81.7 \\
TNT-B~\cite{tnt} & 66.0 & 14.1 & 82.8 \\
Swin-S~\cite{swin} & 50.0 & 8.7 & 83.0 \\
PVTv2-B4~\cite{pvtv2} & 62.6 & 10.1 & 83.6 \\
\rowcolor{gray!20}DiT-B4(ours) & 126.0 & 9.9 & \textbf{84.5} \\
\Xhline{0.5pt}
DeiT-Base/16~\cite{deit}       & 86.6 & 17.6 & 81.8 \\
Swin-B~\cite{swin}             & 88.0 & 15.4 & 83.3 \\
PVTv2-B5~\cite{pvtv2}          & 82.0 & 11.8 & 83.8 \\
\rowcolor{gray!20}DiT-B5(ours) & 158.0 & 10.3 & \textbf{84.8} \\
\end{tabular}}
\end{minipage}
\hfill
\begin{minipage}{0.54\linewidth}
\caption{Comparing with ResNet, Swin Transformer, PVTv2, Focal~\cite{focal} across different object detection methods, ``$AP^b$'' signifies the average precision of bounding box AP. The calculation of ``GFLOPs'' is based on the scale of $1280 \times 800$.}
\label{tab:det_compare}
\centering
\scalebox{0.86}{
\begin{tabular}{l|c|c|ccc}
Backbone & Method & GLOPs & $AP^b$ & $AP^b_{50}$ & $AP^b_{75}$  \\
\Xhline{0.5pt}
ResNet50\cite{resnet} & \multirow{4}*{\makecell{Cascade \\ Mask \\ R-CNN~\cite{Sparsercnn}}} & 739 & 46.3 & 64.3 & 50.5  \\
ConvNeXt-S\cite{convnet} &      & 827 & 51.9 & 70.8 & 56.5 \\
Swin-T\cite{swin}     &         & 745 & 50.5 & 69.3 & 54.9 \\
PVTv2-B2\cite{pvtv2}  &         & 788 & 51.1 & 69.8 & 55.3 \\
\rowcolor{gray!20}DiT-B2(ours)   & & 745 & \textbf{52.3} & \textbf{71.0} & \textbf{56.9} \\
\Xhline{0.5pt}
ResNet50\cite{resnet} & \multirow{4}*{ATSS~\cite{atss}} & 205 & 43.5 & 61.9 & 47.0 \\
Swin-T\cite{swin}     &          & 215 & 47.2 & 66.5 & 51.3 \\
Focal-T\cite{focal}   &          & 239 & 49.5 & 68.8 & 53.9 \\
PVTv2-B2\cite{pvtv2}  &          & 258 & 49.9 & 69.1 & 54.1 \\
\rowcolor{gray!20}DiT-B2(ours)    & & 247 & \textbf{51.2} & \textbf{70.5} & \textbf{55.7} \\
\Xhline{0.5pt}
ResNet50\cite{resnet} & \multirow{4}*{GLF~\cite{gfocalloss}} & 208 & 44.5 & 63.0 & 48.3 \\
Swin-T\cite{swin}     &          & 215 & 47.6 & 66.8 & 51.7 \\
PVTv2-B2-Li\cite{pvtv2}  &       & 197 & 49.2 & 68.2 & 53.7 \\
PVTv2-B2\cite{pvtv2}  &          & 261 & 50.2 & 69.4 & 54.7 \\
\rowcolor{gray!20}DiT-B2(ours) &    & 235 & \textbf{51.3} & \textbf{70.6} & \textbf{55.9} \\
\Xhline{0.5pt}
ResNet50\cite{resnet} & \multirow{4}*{\makecell{Sparse \\R-CNN~\cite{Sparsercnn}}} & 166 & 44.5 & 63.4 & 48.2 \\
Swin-T\cite{swin}     &          & 172 & 47.9 & 67.3 & 52.3 \\
Focal-T\cite{focal}   &          & 196 & 49.0 & 69.1 & 53.2 \\
PVTv2-B2\cite{pvtv2}  &          & 215 & 50.1 & 69.5 & 54.9 \\
\rowcolor{gray!20}DiT-B2(ours)   &  & 202 & \textbf{51.3} & \textbf{70.5} & \textbf{56.0} \\
\end{tabular}}
\end{minipage}
\vspace{-10pt}
\end{table}

\vspace{-3pt}
\subsection{Object detection and instance segmentation}
Our experiments on the challenging COCO benchmark~\cite{coco} involved object detection settings. For training, we used the COCO train2017 dataset (118k images) and for evaluation, we used the val2017 dataset (5k images). We verify the effectiveness of DiT backbones on top of two standard detectors, namely Mask R-CNN~\cite{maskrcnn} and RetinaNet~\cite{retinanet}. We leveraged ImageNet pre-trained weights to initialize backbones alongside Xavier~\cite{Xavier} to initialize newly added layers. To train our detection models, we utilized 8 V100 GPUs with a batch size of 16 and applied AdamW~\cite{adamw} for optimization with an initial learning rate of $1 \times 10^{-4}$. As per common practices~\cite{mmdetection,maskrcnn,iqdet,borderdet}, we employed either a $1 \times$ or $3 \times$ training schedule (i.e., 12 or 36 epochs) respectively. During training, we resized the images to have a shorter side of 800 pixels, while ensuring that the longer side did not exceed 1,333 pixels. For models trained using the $3 \times$ schedule, we randomly resized the shorter side of the input image within the range of [640, 800]. In the testing phase, however, the shorter side of the input image was fixed at 800 pixels.

In order to validate the effectiveness of our DiT, we trained four different object detectors including Cascade R-CNN~\cite{Cascadercnn}, ATSS~\cite{atss}, RepPoints~\cite{gfocalloss} and Sparse R-CNN~\cite{Sparsercnn} by following the approach of PVTv2~\cite{pvtv2}. We use DiT as the backbone for all four models and employed a $1\times$ training schedule. The box mAPs on the COCO validation set are presented in Tab.~\ref{tab:det_compare}. Our DiT-B2 achieved superior performance compared to Swin-Tiny, PVTv2-B2 by 1.2-4 points on all methods. These consistent and substantial improvements across various detection methods, as well as RetinaNet and Mask R-CNN, indicate that our DiT can serve as a versatile backbone for a wide range of object detection applications.

To further verify the effectiveness of our proposed DiT, the performance comparison between our DiT, CNN-based models and current Transformer-based state-of-the-art methods are presented in Tab.~\ref{tab:det3x}, with bbox mAP ($AP^b$) and mask mAP ($AP^m$) reported. we consistently outperform CNN-based models by a margin of 6.7-8.5 points. Furthermore, compared with other methods that utilize multi-scale Transformer architectures, we demonstrate significant improvements across all settings and metrics. Notably, in comparable settings, our DiT achieves a mAP gain of 0.8-1.1 points over the current best approach, PVTv2~\cite{pvtv2} and Focal~\cite{focal}. Unlike other multi-scale Transformer models, our approach enables both low-resolution fine-grain and high-resolution coarse-grain interactions for the visual token. We also provide more comprehensive comparisons by training our models with a 1× schedule and presenting the detailed numbers for RetinaNet and Mask R-CNN in Tab.~\ref{tab:det1x}, along with the number of associated computational costs for each model. Even in $1\times$ schedule, our models demonstrate a gain of 0.9-1.1 over the SoTA models in comparable settings.

\begin{table}[t]
\caption{Object detection and instance segmentation on COCO val2017. ``\#P'' denotes the parameter number. $AP^{b}$, $AP^{m}$ refers to the bounding box AP and mask AP, respectively.}
\label{tab:det3x}
\centering
\setlength{\tabcolsep}{1.2mm}
\scalebox{0.75}{
\begin{tabular}{l|c|ccc|ccc|c|ccc|ccc}
\multirow{2}{*}{Backbone} & \multicolumn{7}{c}{RetinaNet $3\times$ schedule + MS} & \multicolumn{7}{|c}{Mask R-CNN $3\times$ schedule + MS}  \\
\cline{2-15}
~                         & GFLOPs & $AP^{b}$ &$AP_{50}^{b}$&$AP_{75}^{b}$&$AP_{S}^{b}$&$AP_{M}^{b}$&$AP_{L}^{b}$&GFLOPs& $AP^{b}$ &$AP_{50}^{b}$&$AP_{75}^{b}$&$AP^{m}$&$AP_{50}^{m}$&$AP_{75}^{m}$\\
\Xhline{1pt}
\Xhline{0.5pt}
ResNet50\cite{resnet} & 239 & 39.0 & 58.4 & 41.8 & 22.4 & 42.8 & 51.6 & 260 & 41.0 & 61.7 & 44.9 & 37.1 & 58.4 & 40.1 \\
PVT-Small\cite{pvt}   & 226 & 42.2 & 62.7 & 45.0 & 26.2 & 45.2 & 57.2 & 245 & 43.0 & 65.3 & 46.9 & 39.9 & 62.5 & 42.8 \\
Swin-Tiny\cite{swin}  & 245 & 45.0 & 65.9 & 48.4 & 29.7 & 48.9 & 58.1 & 264 & 46.0 & 68.1 & 50.3 & 41.6 & 65.1 & 44.9 \\
ConvNeXt-T\cite{convnet}&243& 45.2 & 65.7 & 48.5 & 28.8 & 49.3 & 58.9 & 262 & 46.2 & 67.9 & 50.8 & 41.7 & 65.0 & 44.9 \\
Focal-Tiny\cite{focal}& 265 & 45.5 & 66.3 & 48.8 & 31.2 & 49.2 & 58.7 & 291 & 47.2 & 69.4 & 51.9 & 42.7 & 66.5 & 45.9 \\
MViTv2-T\cite{MViTv2} & -   & -    & -    & -    & -    & -    & -    & 279 & 48.2 & 70.9 & 53.3 & 43.8 & 67.9 & 47.2 \\
PVTv2-B2~\cite{pvtv2} & 290 & 46.4 & 67.6 & 50.1 & 30.6 & 50.2 & 60.2 & 309 & 47.8 & 69.6 & 52.7 & 42.9 & 66.7 & 46.3 \\
\rowcolor{gray!20}DiT-B2(ours)          & 278 & \textbf{47.5} & \textbf{68.9} & \textbf{51.4} & \textbf{31.6} & \textbf{50.3} & \textbf{61.0} & 290 & \textbf{48.9} & \textbf{70.5} & \textbf{53.8} & \textbf{43.7} & \textbf{67.6} & \textbf{47.2} \\
\Xhline{0.5pt}
ResNet101\cite{resnet}& 315 & 40.9 & 60.1 & 44.0 & 23.7 & 45.0 & 53.8 & 336 & 42.8 & 63.2 & 47.1 & 38.5 & 60.1 & 41.3 \\
PVT-Medium\cite{pvt}  & 283 & 43.2 & 63.8 & 46.1 & 27.3 & 46.3 & 58.9 & 302 & 44.2 & 66.0 & 48.2 & 40.5 & 63.1 & 43.5 \\
Swin-Small\cite{swin} & 335 & 46.4 & 67.0 & 50.1 & 31.0 & 50.1 & 60.3 & 354 & 48.5 & 70.2 & 53.5 & 43.3 & 67.3 & 46.6 \\
Focal-Small\cite{focal}&367 & 47.3 & 67.8 & 51.0 & 31.6 & 50.9 & 61.1 & 401 & 48.8 & 70.5 & 53.6 & 43.8 & 67.7 & 47.2 \\
PVTv2-B3~\cite{pvtv2} & 378 & 46.8 & 68.0 & 50.5 & 31.2 & 50.7 & 60.4 & 397 & 48.1 & 69.9 & 53.1 & 43.2 & 67.0 & 46.6 \\
\rowcolor{gray!20}DiT-B3(ours)      & 355 & \textbf{47.6} & \textbf{68.5} & \textbf{52.0}& \textbf{31.7} & \textbf{51.6} & \textbf{61.1} & 369 & \textbf{48.8} & \textbf{70.7} & \textbf{53.7} & \textbf{43.7} & \textbf{67.7} & \textbf{47.3} \\
\Xhline{0.5pt}
ResNeXt101-64x4d\cite{resnet} & 473 & 41.8 & 61.5 & 44.4 & 25.2 & 45.4 & 54.6 & 493 & 44.4 & 64.9 & 48.8 & 39.7 & 61.9 & 42.6 \\
PVT-Large\cite{pvt}           & 345 & 43.4 & 63.6 & 46.1 & 26.1 & 46.0 & 59.5 & 364 & 44.5 & 66.0 & 48.3 & 40.7 & 63.4 & 43.7 \\
Swin-Base\cite{swin}          & 477 & 45.8 & 66.4 & 49.1 & 29.9 & 49.4 & 60.3 & 496 & 48.5 & 69.8 & 53.2 & 43.4 & 66.8 & 46.9 \\
Focal-Base                    & 514 & 46.9 & 67.8 & 50.3 & \textbf{31.9} & 50.3 & 61.5 & 533 & 49.0 & 70.1 & 53.6 & 43.7 & 67.6 & 47.0 \\
PVTv2-B4~\cite{pvtv2}         & 482 & 47.1 & 67.8 & 50.7 & 30.2 & 51.0 & 62.0 & 500 & 48.5 & 69.2 & 52.9 & 43.1 & 66.6 & 46.8  \\
\rowcolor{gray!20}DiT-B4(ours)& 456 & \textbf{48.0} & \textbf{68.7} & \textbf{51.7} & 31.7 & \textbf{51.9} & \textbf{63.0} & 465 & \textbf{49.4} & \textbf{70.2} & \textbf{53.8} & \textbf{44.0} & \textbf{67.8} & \textbf{47.9} \\
\end{tabular}}
\vspace{-6pt}
\end{table}

\begin{table}[ht]
\begin{minipage}{0.52\linewidth}
\centering
\setlength{\tabcolsep}{1.4mm}
\caption{Comparisons with CNN and Transformer baselines and SoTA methods on COCO object detection. The box mAP ($AP^{b}$) and mask mAP ($AP^{m}$) are reported with 1× schedule.}
\label{tab:det1x}
\scalebox{0.84}{
\begin{tabular}{l|c|cc}
\multirow{2}{*}{Backbone} & RetinaNet & \multicolumn{2}{|c}{Mask R-CNN} \\
\cline{2-4}
~ & $AP^b$ & $AP^b$ & $AP^m$ \\
\Xhline{1pt}
ResNet18~\cite{resnet}           & 31.8 & 34.0 & 31.2 \\
PVTv1-Tiny~\cite{pvt}            & 36.7 & 36.7 & 35.1 \\
PVTv2-B1~\cite{pvtv2}            & 41.2 & 41.8 & 38.8 \\
\rowcolor{gray!20}DiT-B1(ours)   & \textbf{42.3} & \textbf{42.9} & \textbf{39.9} \\
\Xhline{0.6pt} 
ResNet50~\cite{resnet}           & 36.3 & 38.0 & 34.4 \\
PVTv1-Small~\cite{pvt}           & 40.4 & 40.4 & 37.8 \\
Swin-Tiny~\cite{swin}            & 42.0 & 43.7 & 39.8 \\
Focal-Tiny~\cite{focal}          & 43.7 & 44.8 & 41.0 \\
PVTv2-B2~\cite{pvtv2}            & 44.6 & 45.3 & 41.2 \\
\rowcolor{gray!20}DiT-B2(ours)   & \textbf{45.6} & \textbf{46.4} & \textbf{42.2} \\
\Xhline{0.6pt}
ResNet101~\cite{resnet}          & 38.5 & 40.4 & 36.4 \\
PVTv1-Medium~\cite{pvt}          & 41.9 & 42.0 & 39.0 \\
Swin-Small~\cite{swin}           & 45.0 & 46.5 & 42.1 \\
Focal-Small~\cite{focal}         & 45.6 & 47.4 & 42.8 \\
PVTv2-B3~\cite{pvtv2}            & 45.9 & 47.0 & 42.5 \\
\rowcolor{gray!20}DiT-B3(ours)   & \textbf{46.8} & \textbf{48.0} & \textbf{41.7} \\
\Xhline{0.6pt}
PVTv1-Large~\cite{pvt}           & 42.6 & 42.9 & 39.5 \\
Swin-Base~\cite{swin}            & 45.0 & 46.9 & 42.3 \\
Focal-Base~\cite{focal}          & 46.3 & 47.8 & 43.2 \\
PVTv2-B4~\cite{pvtv2}            & 46.1 & 47.5 & 42.7 \\
\rowcolor{gray!20}DiT-B4(ours)   & \textbf{46.9} & \textbf{48.2} & \textbf{43.3} \\
\end{tabular}}
\end{minipage}
\hfill
\begin{minipage}{0.46\linewidth}
\centering
\setlength{\tabcolsep}{1.4mm}
\caption{Semantic segmentation performance of different backbones on the ADE20K validation set. ``GFLOPs'' is calculated under the input scale of $512 \times 512$. We verify the effectiveness of DiT backbones on the Semantic FPN framework which is similar to common practices like PVTv2.}
\label{tab:segment}
\scalebox{0.84}{
\begin{tabular}{l|c|c}
\multirow{2}{*}{Backbone} & \multicolumn{2}{c}{Semantic FPN}  \\
\cline{2-3}
~  & GFLOPs & mIoU(\%) \\ 
\Xhline{1pt}
ResNet18~\cite{resnet} & \textbf{32.2} & 32.9 \\
PVTv1-Tiny~\cite{pvt} & 33.2 & 35.7 \\
PVTv2-B1~\cite{pvtv2} & 34.2 & 42.5 \\
\rowcolor{gray!20}DiT-B1(ours)          & 32.7 & \textbf{45.0} \\
\Xhline{0.6pt}
ResNet50~\cite{resnet}  & 45.6 & 36.7 \\
PVTv1-Small~\cite{pvt}  & 44.5 & 39.8 \\
PVTv2-B2-Li~\cite{pvtv2}  & \textbf{41.0} & 45.1 \\
PVTv2-B2~\cite{pvtv2}     & 45.8 & 45.2 \\
\rowcolor{gray!20}DiT-B2(ours) & 43.4 & \textbf{47.4} \\
\Xhline{0.6pt}
ResNet101~\cite{resnet}   & 65.1 & 38.8 \\
ResNeXt101-32x4d~\cite{resnet} & 64.7 & 39.7 \\
PVTv1-Medium~\cite{pvt}   & 61.0 & 41.6 \\
PVTv2-B3    ~\cite{pvtv2} & 62.4 & 47.3 \\
\rowcolor{gray!20}DiT-B3(ours)     & \textbf{59.0 } & \textbf{48.7} \\
\Xhline{0.6pt}
ResNeXt101-64x4d\cite{resnet} &103.9 & 40.2 \\
PVTv1-Large  ~\cite{pvt}      & 79.6 & 42.1 \\
PVTv2-B4     ~\cite{pvtv2}    & 81.3 & 47.9 \\
\rowcolor{gray!20}DiT-B4(ours)     & \textbf{78.4 } & \textbf{49.0} \\
\end{tabular}}
\end{minipage}
\vspace{-6pt}
\end{table}

\subsection{Semantic segmentation}
\vspace{-3pt}
To evaluate the performance of semantic segmentation, We utilized ADE20K~\cite{ade20k} as our benchmark dataset for semantic segmentation, following the methodology in PVTv1~\cite{pvt}. To ensure fair comparisons, we evaluated the performance of DiT using the Semantic FPN~\cite{Semantic_FPN} framework. During the training phase, we initialize the backbone with pre-trained weights from ImageNet~\cite{imagenet} and apply Xavier~\cite{Xavier} initialization to the newly added layers. Our model is optimized with AdamW~\cite{adamw} using an initial learning rate of 1e-4. For consistency with common practices~\cite{Semantic_FPN,deeplab}, we train our models for 40k iterations with a batch size of 16 on 4 V100 GPUs. A polynomial decay schedule with a power of 0.9 is employed to decay the learning rate. We randomly resize and crop the image to 512×512 during training, while only the shorter side is rescaled to 512 pixels during testing.

According to Tab.~\ref{tab:segment}, DiT outperforms PVTv2 and other models in semantic segmentation. The results show that DiT-B1/B2/B3/B4 achieve 1.1-2.5 higher performance than other method, with similar parameters and GFLOPs. Additionally, although the GFLOPs of DiT-v2 are 6\% lower than those of SoTA PVTv2, the mIoU is still 2.2 points higher (47.4 vs 45.2). It demonstrates that DiT backbones extract powerful features for semantic segmentation, benefiting from our dynamic routing selection strategy.

\begin{table}[h]
\setlength{\tabcolsep}{0.33mm}
\begin{minipage}{0.33\linewidth}
\centering
\caption{``Static'' denotes that we initialize the parameters of the gating module and use gating masks to select the routing path. We fix all the gating parameters and set them to 1.0 which is marked as ``fully'' connected.}
\label{tab:ablation1}
\scalebox{0.86}{
\begin{tabular}{l|ll}
\toprule
Model & GFLOPs & Acc.(\%) \\
\Xhline{0.5pt}
Static   & 3.8                                      & 80.7\textcolor{darkgreen}{(-2.4)} \\
Dynamic(ours)  & 3.8                                      & 83.1 \\
Fully    & 15.6\textcolor{darkgreen}{($\times4.1$)} & 83.5\textcolor{darkgreen}{(+0.4)} \\
\bottomrule
\end{tabular}}
\end{minipage}
\hfill
\begin{minipage}{0.33\linewidth}
\centering
\caption{Different gating mask generation methods consist of \textit{random} mask generation, selection by \textit{attention} probabilities and selection by a \textit{learnable} routing gating module.}
\label{tab:ablation2}
\scalebox{0.88}{
\begin{tabular}{l|cl}
\toprule
Model & GFLOPs & Acc.(\%) \\
\Xhline{0.5pt}
Random & 3.8 & 79.4\textcolor{darkgreen}{(-3.7)}\\
Attention & 3.8 & 81.3\textcolor{darkgreen}{(-1.8)} \\
Learnable(ours) & 3.8 & 83.1 \\
\bottomrule
\end{tabular}}
\end{minipage}
\hfill
\begin{minipage}{0.31\linewidth}
\centering
\caption{Ablation studies of each contribution, including dynamic scale and depth comparison on the ImageNet validation set. The baseline of them is a pure PVTv2\cite{pvtv2}. }
\label{tab:ablation3}
\scalebox{0.86}{
\begin{tabular}{cc|cc}
\toprule
\makecell{Scale} & \makecell{Depth} & GFLOPs & Acc.(\%) \\
\Xhline{1pt}
           &            & 4.0 & 82.0 \\
\checkmark &            & 8.4 & 82.9 \\
           & \checkmark & 3.7 & 82.2 \\
\checkmark & \checkmark & 3.8 & 83.1 \\
\bottomrule
\end{tabular}}
\end{minipage}
\vspace{-10pt}
\end{table}

\subsection{Ablation studies}
\vspace{-3pt}
The studies involve comparing various routing selection methods, including static, dynamic, and fully connected token routing as depicted in Tab.~\ref{tab:ablation1}. We see our method achieves better trade-offs between Flops and accuracy than other routing selection methods. The fully-connected strategy significantly increases computational complexity and leads to minor improvement(+0.4). We also compare our method with other gating mask generation strategies including ``random'' mask generation, selection by tokens ``attention'' probabilities, and our ``learnable'' gating mask generation in Tab.~\ref{tab:ablation2}. We find that our data-dependent method chooses better token routing paths and balances the complexity of the network. As illustrated in Tab.~\ref{tab:ablation3}, the ablation studies of the dynamic scale and dynamic network depth denote our methods lead to better performance without introducing computational complexity.

\subsection{Analysis}
\vspace{-3pt}
\paragraph{Visualization of the routing selection.} 

\begin{figure}[t]
\vspace{-10pt}
\centering
\includegraphics[clip=true, width=1.0\linewidth]{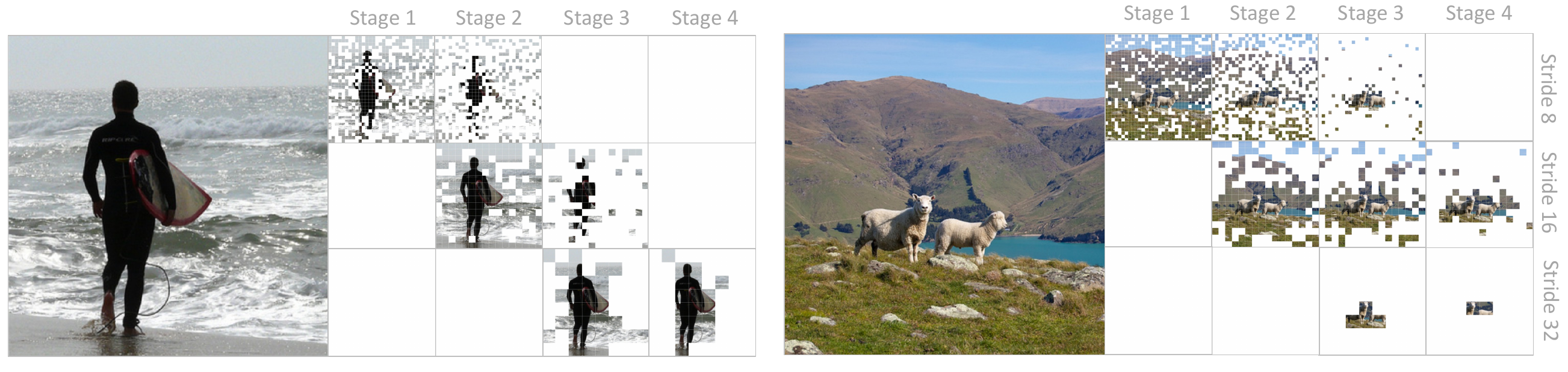}
\caption{Visualization of the dynamic tokens routing. We show the original input image and the routing results of the four stages, where each row of the images denotes the multi-scale feature map scaling from stride of 8 to 32. The masks(in white) represent the corresponding tokens that select the paths of identify mapping. Our method can gradually focus on the most representative regions in the image and the feature map(stride=32) is more sensitive to large-scale objects. The observation implies that DynamicViT exhibits superior interpretability. }
\label{fig:visualization}
\end{figure}

To further investigate the behavior of DiT, we visualize the routing selection procedure in Fig.~\ref{fig:visualization}. We show the original input image and the token routing selection results after the four stages and multi-scale feature
map scaling from stride of 8 to 32, where the masks in white represent the corresponding tokens select the paths of identify mapping.

We find that through the token paths selection based on the dimension of scale and depth, our DiT may neglect the uninformative tokens(e.g. background) and finally focus on the objects in the images. For the low-resolution feature(e.g. stride=32), DiT is more sensitive to large objects. This phenomenon also suggests that the DiT leads to better interpretability, i.e., it can locate the important parts in the image which contribute most to the final output. And it also illustrates that DiT may adaptively select the routing path according to the spatial scale of the objects and the difficulty of the recognition.

\begin{figure}[h]
\vspace{-10pt}
\centering
\includegraphics[clip=true, width=0.8\linewidth]{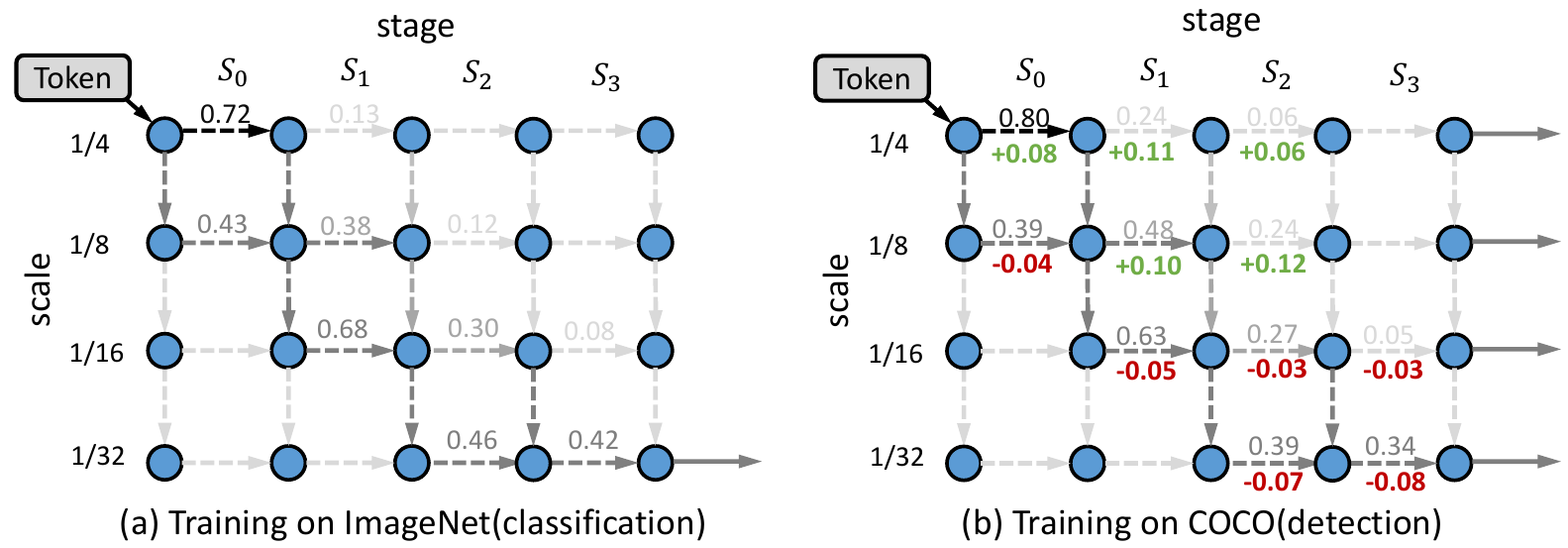}
\caption{Statistical Analysis of the dynamic routing selection of ImageNet(Cls) and COCO(Det). The numbers over the line represent the ratio of tokens being skipped.}
\label{fig:line_weight}
\end{figure}

\paragraph{Statistical Analysis of the routing selection.} 
Besides the sample-wise visualization we have shown above, we are also interested in the statistical characteristics of the routing selection, i.e., what kind of general patterns does the DiT learn from the different datasets? We then use the DiT to generate the gating mask for all the images in the ImageNet set and COCO detection set to compute the probability of mask in different scales and stages, as shown in Fig.~\ref{fig:line_weight}. Compared to the classification task(ImageNet), the spatial scales of objects in the detection task(COCO) are smaller. Unsurprisingly, we find the token selection of the high-resolution features tends to be calculated.


\section{Conclusion}
\vspace{-4pt}
In this work, we present the dynamic routing for dense prediction. The key difference from prior works lies in that we generate data-dependent dynamic routing paths according to the scale and difficulty distribution of each token. The \textit{dynamic routing gate} selects the scale transformation and calculation routes in an end-to-end manner, which learns to skip some useless operations for the trade-off between accuracy and Flops. Gumbel-Softmax and \textit{dynamic routing gate} techniques are also incorporated for the end-to-end training of the transformer model together with the prediction module. Through inserting the dynamic token routing into the backbone, each token will be calculated on an adaptive spatial scale and the dynamic number of transformer layers. DiT achieves state-of-the-art performance on COCO object detection and ADE20K semantic segmentation, surpassing previous best methods. However, we believe our dynamic token routing is also applicable to monolithic vision Transformers and Transformers in both vision and language domains. We leave this as a promising direction to further explore in the future.

{\small
\bibliographystyle{unsrt}
\bibliography{ref.bib}
}




\medskip

\end{document}